\documentclass[letterpaper, 10 pt, conference]{ieeeconf}  

\IEEEoverridecommandlockouts                              

\usepackage{graphicx} 
\usepackage{amsmath} 
\usepackage{hyperref}  
\usepackage{xurl}      
\usepackage[dvipsnames]{xcolor}
\usepackage{threeparttable}
\usepackage{makecell}

\title{\LARGE \bf
Seeing Through Occlusion: Deterministic Arm Kinematic Correction for Robot Teleoperation
}

\author{
Thomas M. Kwok$^{1}$, Nicholas Koenig$^{1}$, Yue Hu$^{1}$
\thanks{$^{1}$ Department of Mechanical and Mechatronics Engineering, University of Waterloo, Canada {\tt\footnotesize nekoenig@uwaterloo.ca, thomasm.kwok@uwaterloo.ca, yue.hu@uwaterloo.ca}}
\thanks{This work was supported
in part by the National Research Council (NRC) under Grant NRC-AiP-302-1; and the support of Waterloo RoboHub for the equipment used. We acknowledge the support of the Natural Sciences and Engineering Research Council of Canada (NSERC), funding reference number RGPIN-2022-03857.}
}

\begin{document}

\maketitle
\thispagestyle{empty}
\pagestyle{empty}

\begin{abstract}

Markerless, single-RGB-D-camera motion capture provides a low-cost and non-invasive alternative to conventional marker-based systems for robot teleoperation; however, depth estimation often degrades in the presence of self-occlusion, particularly during upper-limb motion. This paper presents an Arm Kinematic Correction (AKC) method that improves depth estimation by enforcing geometric constraints based on constant arm lengths. The proposed approach reconstructs occluded joint depths by leveraging wrist positions and predefined arm lengths via a deterministic formulation based on the Pythagorean theorem, thereby avoiding the need for complex probabilistic modeling or parameter tuning. Experimental validation against a Vicon reference system demonstrates reliable performance for both static and dynamic joint motions, evaluated using root-mean-square error (RMSE) and Pearson correlation. Furthermore, motion-mapping teleoperation is successfully demonstrated in both simulated and physical robot environments. The results show that AKC enhances robustness and preserves anatomical consistency under long-duration, severe self-occlusion, even when paired with less reliable temporal filters, highlighting its practicality for real-time applications such as robot teleoperation and human–robot interaction.

\end{abstract}

\section{INTRODUCTION}

The global population is aging rapidly \cite{prb_ageing}. According to the World Health Organization \cite{noauthor_ageing_nodate}, by 2050, individuals aged 60 and above will account for 22\% of the global population. This demographic shift is expected to place increasing pressure on caregiving systems, particularly in supporting older adults with daily activities such as food preparation. At the same time, the limited availability of professional caregivers, combined with the geographically dispersed population in countries such as the United States and Canada, further increases the physical demands on caregivers.

Remote robot-assisted caregiving has emerged as a promising approach to alleviating these challenges, enabling caregivers to support older adults via teleoperated robots \cite{kwok_leveraging_2025, lyu_teleoperation_2020, shah_toward_2024}. Many assistive tasks rely on arm manipulation, such as feeding and housekeeping \cite{petrich_2022}, making reliable arm motion capture and motion-mapping teleoperation essential for remote care.

To capture arm motion, various teleoperation interfaces have been explored, including wearable inertial measurement units (IMUs) \cite{vskulj2021wearable} and virtual reality (VR)-based systems \cite{robotics12060163}. However, IMU-based approaches are susceptible to drift over time \cite{kwok_2023}, while VR systems may induce motion sickness \cite{9133071} and require continuous use of head-mounted devices. In addition, these systems often rely on wired or body-mounted hardware, which may restrict user mobility and increase setup complexity. In contrast, vision-based motion capture provides a minimally invasive or non-invasive alternative that does not require wearable sensors, making it well-suited for practical teleoperation scenarios.

Motion capture (mocap) systems based on markers \cite{kakavand_2025, federolf_2025} or multi-camera setups \cite{moon_multiple_2016, chromy_2025, kakavand_2025, federolf_2025} typically require complex configurations, including large capture volumes, precise marker placement, and manual calibration of both camera and participant parameters. These requirements can limit their feasibility in clinical environments, particularly in emergency situations \cite{Castro_2026} involving older adults, where caregivers may need to remotely operate robotic systems within a short time frame without lengthy calibration procedures. In addition, such systems are difficult to accommodate in typical clinic offices, where available space is limited.

To address these limitations, a single RGB-D camera–based mocap offers a more practical alternative due to its low cost, non-invasiveness, and relatively simple setup. It enables human pose estimation in confined environments and has been increasingly adopted to support motion-mapping teleoperation by estimating the operator's physical state \cite{bin_survey_2025}, thereby supporting teleoperation ~\cite{kwok_leveraging_2025}. However, single RGB-D mocap systems for teleoperation still face two persistent challenges that remain difficult to address in practice.

\begin{figure} [htbp]
    \centering
    \includegraphics[width=7.5cm, height=6.5cm]{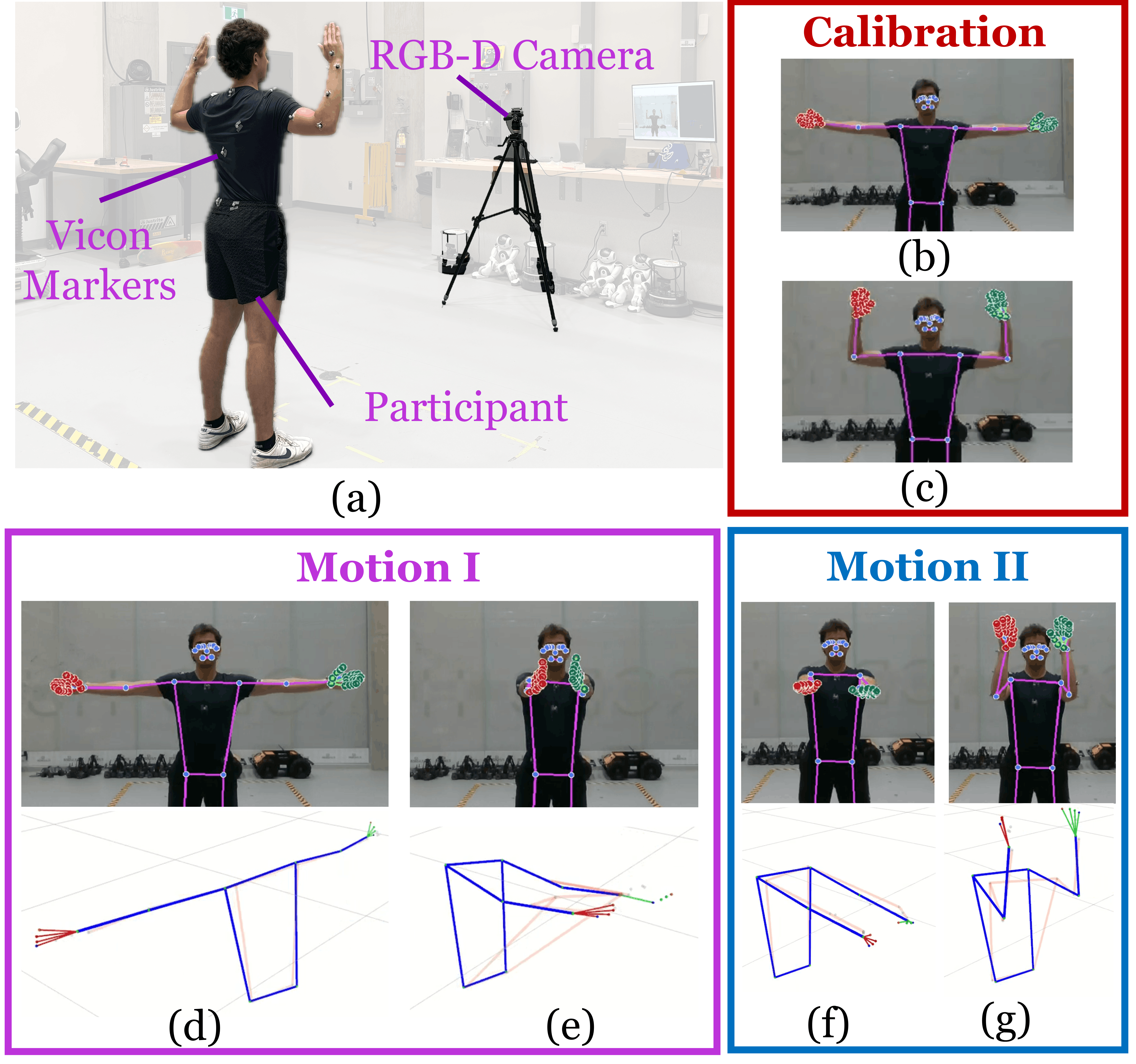}
    \caption{ (a) Experiment Setup.(b) T-pose, (c) L-pose for calibration. During Motion I, the participant performed (d) T-pose, (e) horizontal adduction, and back to T-pose. For Motion II, he performed (c) arm extension, (d) elbow flexion, and back to arm extension. In (d)-(g), (top) 2D Mediapipe output, and (bottom) 3D human skeleton corrected with the AKC. The blue skeletons represent the AKC output, while the red are raw camera input.}
    \label{fig:poses_exp}
\end{figure}

First, depth estimation becomes unreliable under self-occlusion \cite{zhou_review_2024}. As shown in Fig.~\ref{fig:poses_exp} (e-g), distal joints (e.g., hand and wrist) occlude proximal joints (e.g., elbow and shoulder), leading to incorrect depth assignments. Such errors introduce instability in teleoperation, causing abrupt and unintended motion of the robot.

\begin{figure} [htbp]
    \centering
    \includegraphics[width=4.5cm]{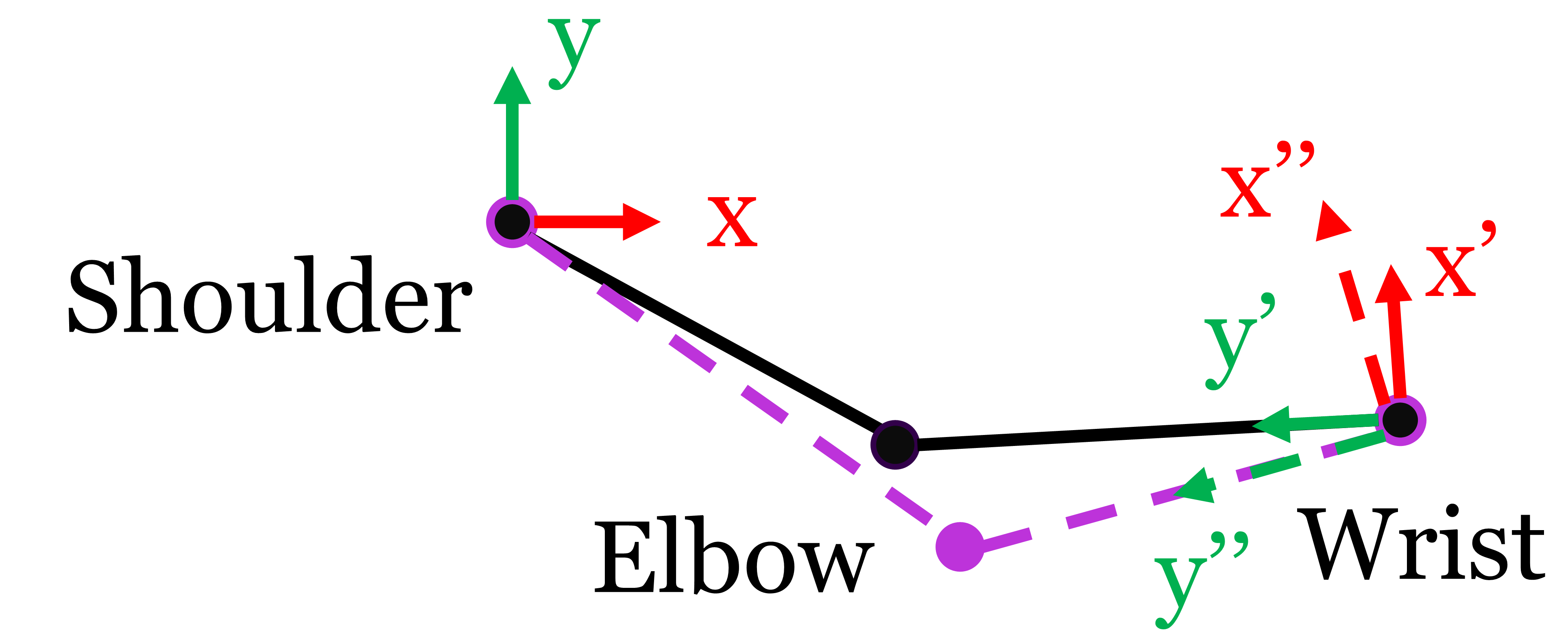}
    \caption{Effect of inconsistent arm length on control frame construction. The figure compares frames generated under constant arm length and inconsistent length, showing distortion in the resulting wrist frame (purple). $(x, y)$ denotes the shoulder base frame, $(x', y')$ the reference wrist frame, and $(x'', y'')$ the distorted frame.}
    \label{fig:inconsistent_arm}
\end{figure}

Second, estimated poses often exhibit inconsistent arm lengths, violating anatomical constraints and degrading motion fidelity. While filtering methods such as the Kalman Filter (KF) \cite{kalman_1960, moon_multiple_2016, kf_paper} improve temporal smoothness, they process landmarks independently and do not enforce kinematic constraints between joints. This results in inconsistent joint configurations, leading to errors in control frame construction (Fig.~\ref{fig:inconsistent_arm}) and reduced teleoperation reliability.

Recent advancements in depth sensor-based motion capture (D-Mocap) aimed to improve robustness under occlusion and noise \cite{zhou_review_2024}. Probabilistic approaches, such as particle filters \cite{tripathy_constrained_2018}, incorporate motion models to smooth noisy observations, while variants such as the Tobit Kalman Filter \cite{lannan_human_2022} explicitly address occlusion. The Extended Kalman Filter (EKF) \cite{welch_1995, joukov_2018, kwok_2023}, optimization-based methods \cite{tang_1999, cotton_2023}, and evolutionary algorithms \cite{das_improvement_2017} further enforce kinematic constraints during pose reconstruction. In addition, learning-based approaches \cite{wang_spatio_temporal_2021} improve data fidelity by suppressing implausible motions, and hybrid frameworks that fuse kinematic models with deep learning and filtering techniques \cite{pham_2025} have demonstrated improved robustness.

Despite these advances, many existing methods rely on complex model design, parameter tuning, or iterative optimization, which can limit their efficiency and robustness. In particular, severe and prolonged self-occlusion remains challenging, often resulting in unstable depth estimation and inconsistencies in anatomical structure.

In this paper, we present a human pose estimation method using a single RGB-D camera. Our approach employs an arm kinematic correction (AKC) algorithm as an additional layer to correct unreliable poses resulting from self-occlusion and anatomical inconsistencies after the standard pose estimation method. Utilizing the Pythagorean theorem and geometric constraints of human anatomy, it corrects unreliable depth estimates of occluded joints and ensures anatomically coherent poses under self-occlusion. 

The main contributions of this paper are:
\begin{itemize}
    \item A post-processing correction framework for a single RGB-D camera mocap system that improves depth estimation of occluded joints using geometric constraints;
    \item A deterministic geometric approach that enforces constant arm lengths without requiring complex probabilistic modeling or parameter tuning;
    \item Experimental validation demonstrating improved robustness and anatomical consistency for motion-mapping teleoperation.
\end{itemize}

The rest of the paper is organized as follows. Section \ref{sec:method} describes the proposed motion-mapping teleoperation framework using AKC. Section \ref{sec:experiment} presents the experimental validation of AKC and its teleoperation framework. Finally, Section \ref{sec:discussion} and Section \ref{sec:conclusion} include discussion and conclusion.

\section{Motion Capture with Arm Kinematics Correction (AKC)}
\label{sec:method}
Human pose estimation using a single RGB-D camera is commonly achieved with open-source pose landmark detection methods such as OpenPose \cite{openpose_2019} and MediaPipe Pose \cite{mediapipe_pose_landmarker}. In this work, we adopt MediaPipe’s BlazePose model \cite{bazarevsky_blazepose_2020} for 2D landmark detection due to its ease of integration and real-time performance. Depth information is obtained from stereo-based RGB-D sensors such as the Intel RealSense D435 \cite{realsense_depth_guide}, enabling a low-cost markerless motion capture system.

However, such methods do not account for the challenges of teleoperation in single-camera scenarios. As discussed in Section \ref{sec:introduction}, depth estimation becomes unreliable under self-occlusion, where distal joints occlude proximal joints, leading to incorrect depth assignments. The resulting pose estimates may violate anatomical constraints, such as constant arm lengths, producing kinematically inconsistent motions. Further, MediaPipe Pose may fail to detect landmarks due to poor video quality or joints moving outside the camera’s field of view \cite{kwok_leveraging_2025}, resulting in missing $x$, $y$, or $z$ coordinates.

Standard filters improve temporal smoothness by reducing measurement noise and predicting the missing coordinates. Yet, these methods treat each joint independently and do not enforce structural constraints on arm geometry, resulting in inconsistent arm lengths and degraded motion fidelity.

To address these limitations, we introduce the AKC algorithm, a lightweight post-processing module applied to filtered pose landmarks. The proposed method enforces constant upper- and lower-arm lengths by leveraging geometric relationships derived from the Pythagorean theorem. By correcting unreliable depth estimates for occluded joints while maintaining anatomical consistency, AKC produces stable, physically plausible pose estimates suitable for teleoperation.

\subsection{Framework Overview}

\begin{figure} [htbp]
    \centering
    \includegraphics[width=8.7cm,height=4.2cm]{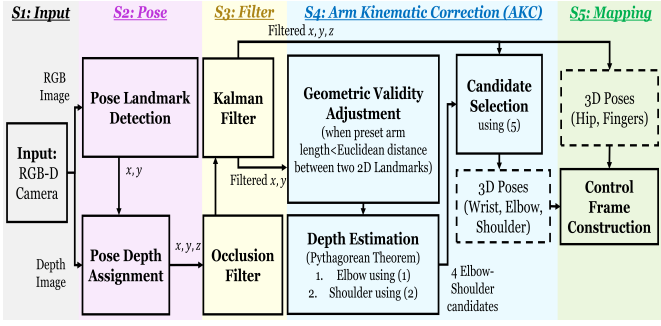}
    \caption{Motion-mapping teleoperation framework with the proposed Arm Kinematic Correction (AKC) algorithm. The pipeline consists of five stages: input acquisition (S1), pose landmark detection (S2), filtering (S3), AKC (S4), and motion mapping (S5).}
    \label{fig:mocap_framework}
\end{figure}

As shown in Fig.~\ref{fig:mocap_framework} and the attached video (00:06–00:24), the proposed framework comprises a multi-stage pipeline and teleoperation using a single RGB-D camera.

In Stage 1, the RGB-D camera captures video with synchronized RGB and depth streams. The RGB input is then processed in Stage 2 (Sec. \ref{subsec:stage2}) by a pose landmark detection algorithm to extract 2D landmarks, which are augmented with depth values to reconstruct the initial 3D pose. 

The resulting pose data are refined in Stage 3 (Sec. \ref{subsec:stage3}) through occlusion handling and temporal filtering: an occlusion filter removes occluded landmarks, while a KF smooths the trajectories and estimates the removed data. 

Building on this, Stage 4 (Sec. \ref{subsec:stage4}) introduces the proposed AKC algorithm as the core component, which enforces anatomically consistent arm motion to mitigate errors arising from self-occlusion and arm-length inconsistencies. Finally, the corrected 3D pose is used in Stage 5 (Sec. \ref{subsec:stage5}) to construct control frames for motion-mapping teleoperation.

\subsection{Stage 2: Pose Landmark Construction}
\label{subsec:stage2}

This stage constructs initial 3D human pose landmarks from RGB-D input and refines them through occlusion handling and temporal filtering. Pose landmarks, including hand, body, and face landmarks, are first detected from RGB input using the MediaPipe Holistic Landmarker \cite{mediapipe_holistic}. This provides reliable 2D coordinates ($x$, $y$) for the selected pose landmarks, including the shoulder, elbow, wrist, hip, index finger metacarpophalangeal (MCP), and pinky MCP. Depth values ($z$) are then extracted from the corresponding depth image and combined with the 2D coordinates via inverse projection under the pinhole camera model to reconstruct initial 3D coordinates.

While the 2D landmark positions are generally robust under self-occlusion, depth estimates are often unreliable for occluded joints, particularly the elbow and shoulder. In addition, some landmarks may be missing due to low visibility or detection failure, resulting in incomplete pose information.

\subsection{Stage 3: Temporal Landmark Filtering}
\label{subsec:stage3}

To improve data quality, an occlusion filter is first applied to discard unreliable measurements. Landmarks with visibility scores below a predefined threshold (e.g., 0.7) are rejected, as low values typically indicate occlusion, motion blur, or poor lighting \cite{mediapipe_pose_landmarker}. 

However, in certain self-occlusion scenarios, the visibility score predicted by BlazePose’s Convolutional Neural Network (CNN) model \cite{bazarevsky_blazepose_2020} may fail to detect corrupted measurements. Thus, a spatial proximity check is introduced: the depth of a proximal joint (e.g., elbow or shoulder) is discarded when its projected 2D position lies within a small neighborhood (e.g., 5\% of the image width) of a distal joint (e.g., wrist), indicating likely occlusion.

After filtering, all landmarks—including those that are missing or discarded—are processed using a KF with a constant-velocity model to smooth trajectories and estimate missing values. The state vector is defined as:

\vspace{-3mm}
\begin{equation}
\mathbf{x}_k =
\begin{bmatrix}
p_k & \dot{p}_k
\end{bmatrix}^T, 
p_k = \begin{bmatrix}x_k & y_k & z_k
\end{bmatrix}^T
\end{equation}
\vspace{-5mm}

where $\mathbf{x}_k$ represents the position and velocity at time step $k$. The system and measurement model are:

\vspace{-3mm}
\begin{equation}
\mathbf{x}_k = \mathbf{A} \mathbf{x}_{k-1} + \mathbf{w}_k, 
\mathbf{z}_k = \mathbf{H} \mathbf{x}_k + \mathbf{v}_k
\end{equation}
\vspace{-5mm}

where $\mathbf{z}_k = p_k^{\mathrm{meas}}$ is the observed position. The measurement noise covariance is defined as:

\vspace{-3mm}
\begin{equation}
\mathbf{R} = \mathrm{diag}(\sigma_x^2, \sigma_y^2, \sigma_z^2), \quad \sigma_z \gg \sigma_x, \sigma_y
\end{equation}
\vspace{-5mm}

to reflect higher uncertainty in depth measurements. When depth observations are unavailable, a partial update is performed using only the $x$ and $y$ components. While the KF provides effective temporal smoothing, more advanced filtering techniques can be incorporated to improve the accuracy. 

To illustrate this flexibility, we implement an EKF with a constrained constant-velocity model, and evaluate its performance in Section~\ref{subsec:filter}. For two connected joints ${p}_i$ and ${p}_j$, the constraint is defined as:

\vspace{-3mm}
\begin{equation}
h(\mathbf{x}) = \|{p}_i - {p}_j\|
\end{equation}
\vspace{-5mm}

which is enforced to match a predefined arm length $l$. The linearized Jacobian is given by:

\vspace{-3mm}
\begin{equation}
\mathbf{H}_c =
\begin{bmatrix}
\frac{(x_i-x_j)}{\|{p}_i - {p}_j\|} & \frac{(y_i-y_j)}{\|{p}_i - {p}_j\|} & \frac{(z_i-z_j)}{\|{p}_i - {p}_j\|} & 0 & 0 & 0
\end{bmatrix}
\end{equation}
\vspace{-5mm}

The EKF update is:

\vspace{-3mm}
\begin{equation}
\mathbf{x} \leftarrow \mathbf{x} + \mathbf{K}_c (l - h(\mathbf{x}))
\end{equation}
\vspace{-5mm}

where $\mathbf{K}_c$ is the Kalman gain using constraint covariance. This acts as a soft constraint to maintain constant arm lengths.

\subsection{Stage 4: Arm Kinematic Correction (AKC)}
\label{subsec:stage4}

The primary objective of the AKC is to replace the filtered pose landmarks with geometrically corrected poses that satisfy anatomical constraints. The KF-filtered landmarks serve as reference estimates, guiding the selection of the most plausible corrected pose.

AKC is based on the assumption that upper- and lower-arm lengths remain constant during motion and that the wrist is less prone to occlusion due to its proximity to the camera. By leveraging this geometric prior, the method reconstructs physically plausible joint configurations from incomplete or corrupted depth measurements. The correction process consists of (i) geometric reconstruction using the Pythagorean theorem and (ii) candidate selection based on spatial, temporal, and anatomical criteria.
    
    \subsubsection{Depth Estimation using the Pythagorean Theorem}
    Given reliable 2D coordinates ($x$, $y$) from pose detection, the depth ($z$) of occluded joints can be reconstructed using geometric constraints. Let $l_f$ and $l_u$ denote the forearm and upper arm lengths, respectively. The depth of the elbow ($p_{e,z}$) and shoulder ($p_{s,z}$) relative to the wrist position ($p_w$) is estimated as:
    
    \vspace{-5mm}
    \begin{equation}
    p_{e,z} = p_{w,z} \pm \sqrt{l_f^2 - (p_{e,x}-p_{w,x})^2 - (p_{e,y}-p_{w,y})^2}
    \label{eqn:pyth_elb}
    \end{equation}
    
    \vspace{-5mm}
    
    \begin{equation}
    p_{s,z} = p_{e,z} \pm \sqrt{l_u^2 - (p_{s,x}-p_{e,x})^2 - (p_{s,y}-p_{e,y})^2}
    \end{equation}
    \label{eqn:pyth_sho}
    \vspace{-5mm}
    
    This formulation yields multiple feasible solutions due to the $\pm$ terms. Specifically, two elbow candidates are generated, and for each elbow candidate, two corresponding shoulder candidates are obtained, resulting in four possible arm configurations. These candidates represent different valid 3D interpretations consistent with the observed 2D geometry and predefined arm lengths.

    However, measurement noise or preceding filtering may occasionally produce negative radicands, leading to invalid (imaginary) solutions. To address this, a geometric validity adjustment is introduced. When such cases occur, the joint position is projected onto the feasible surface by scaling the Euclidean distance between landmarks to match the predefined arm length:

    \vspace{-5mm}
    \begin{equation}
    p_e = p_w + \frac{p_e - p_w}{\|p_e - p_w\|} \, l_f, p_s = p_e + \frac{p_s - p_e}{\|p_s - p_e\|} \, l_u
    \end{equation}
    \label{eqn:arm_ratio}
    \vspace{-3mm}
    
    This adjustment ensures physically valid joint configurations while minimally altering the original pose structure. Since it modifies the measured landmark positions, it is applied only as a fallback mechanism when necessary.
    
    \subsubsection{Candidate Selection}
    \label{subsec:metric}
    
    To identify the most plausible joint configuration, a cost function $c$ is defined to evaluate each candidate pair. For each candidate $i \in \{1, \dots, 4\}$, the total cost is computed as:

    \vspace{-5mm}
    \begin{equation}
    c_i = \sum_{j=1}^{5} w_j m_j
    \end{equation}
    \vspace{-3mm}
    
    where $w_j$ are weighting coefficients and $m_j$ are evaluation metrics. The candidate with the lowest cost is selected as the final estimate.
    
    The evaluation metrics are defined as follows:
    
    \begin{enumerate}
    
    \item \textbf{Elbow spatial consistency:}
    
    \vspace{-3mm}
    \begin{equation}
    m_1 = \| p_e - \hat{p}_e \|
    \end{equation}
    \vspace{-3mm}
    
    Penalizes deviation from the KF-filtered elbow.

    \item \textbf{Shoulder spatial consistency:}
    
    \vspace{-3mm}
    \begin{equation}
    m_2 = \| p_s - \hat{p}_s \|
    \end{equation}
    \vspace{-3mm}
    
    Penalizes deviation from the KF-filtered shoulder.
    
    \item \textbf{Temporal consistency:}

    \vspace{-3mm}
    \begin{equation}
    m_3 = \| p_e - p_e^{prev} \| + \| p_s - p_s^{prev} \|
    \end{equation}
    \vspace{-3mm}
    
    Penalizes abrupt changes from the previous joint positions ($p^{prev}$) to reduce temporal jitter.
    
    \item \textbf{Shoulder width regularization:}

    \vspace{-3mm}
    \begin{equation}
    m_4 = \left| l_s - \| p_{s,l} - p_{s,r} \| \right|
    \end{equation}
    \vspace{-3mm}
    
    Encourages anatomically consistent shoulder width.
    
    \item \textbf{Anatomical plausibility:}
    
    \vspace{-3mm}
    \begin{equation}
    m_5 =
    \begin{cases}
    k, & \text{if } \theta \notin [40^\circ, 180^\circ] \\
    0, & \text{otherwise}
    \end{cases}
    \end{equation}
    \vspace{-3mm}
    
    Penalizes joint configurations outside physiological limits (e.g., $\theta <40^\circ$ and $\theta > 180^\circ$).
    
    \end{enumerate}
    
    This selection strategy balances measurement fidelity, temporal smoothness, and anatomical feasibility, enabling robust reconstruction under self-occlusion.

    Furthermore, selecting optimal candidates becomes challenging when the solution lies near the zero-depth plane. In particular, when the preset arm length exceeds the observed arm length due to measurement noise and landmark detection errors, the radicands in Eqn.~\ref{eqn:pyth_elb} and ~\ref{eqn:pyth_sho} may remain positive even when the arm is fully extended near the zero-depth plane. In such cases, the elbow and shoulder depth become unstable between the $\pm$ solutions.

    To improve robustness, the preset arm length is reduced during candidate search using a shrink factor ($s$), such that the adjusted length is $l_j' = s \, l_j$, where $j \in \{f, u\}$. The effect of this strategy and the associated evaluation metrics will be further analyzed in Section~\ref{subsec:metric_effect}.

\subsection{Comparison: Constrained Optimization (COIK)}
To evaluate the effectiveness of the proposed Arm Kinematic Correction (AKC) algorithm, we implement a constrained optimization-based inverse kinematics (COIK) method as a baseline. This approach reconstructs 3D upper-limb joint positions by enforcing anatomical constraints through optimization.

The method estimates joint positions by minimizing the deviation from filtered landmark measurements while maintaining temporal smoothness. The optimization problem is formulated as:

\vspace{-3mm}
\begin{equation}
\begin{aligned}
\min_{\mathbf{p}} \quad & f(\mathbf{p}) = \|\mathbf{p} - \hat{\mathbf{p}}\|^2 
+ \lambda \|\mathbf{p} - \mathbf{p}^{\,\text{prev}}\|^2 \\
\text{s.t.} \quad 
& \|p_{e,j} - p_{w,j}\|^2 = l_{f,j}^2, \\
& \|p_{s,j} - p_{e,j}\|^2 = l_{u,j}^2, \\
& \theta_{\min} \le \theta(p_{s,j}, p_{e,j}, p_{w,j}) \le \theta_{\max}, \\
&\|p_{s,l} - p_{s,r}\|^2 = l_s^2, 
\end{aligned}
\end{equation}
\label{eqn:coik}
\vspace{-3mm}

where $j \in \{l,r\}$. $\mathbf{p} = [p_{s,l}, p_{e,l}, p_{w,l}, p_{s,r}, p_{e,r}, p_{w,r}]$ contains all joint positions, including shoulder ($s$), elbow ($e$), wrist ($w$) for left ($l$) and right arms ($r$). $\hat{\mathbf{p}}$ denotes the KF-filtered reference, and $\mathbf{p}^{\text{prev}}$ corresponds to the joint positions from the previous frame. The constraints enforce constant forearm length ($l_f$), upper arm length ($l_u$), shoulder width ($l_s$), and anatomically plausible elbow joint angles ($\theta$).

This formulation explicitly enforces kinematic consistency through equality constraints and temporal regularization. However, solving the optimization problem requires iterative computation and careful tuning of the weight ($\lambda$). 

The COIK method serves as a representative baseline to compare with the proposed AKC algorithm, which instead applies a deterministic geometric correction without iterative optimization. A quantitative comparison between AKC and COIK is presented in Sections~\ref{subsec:perform_occlusion}, ~\ref{subsec:arm length}, and \ref{subsec:teleop}.

\subsection{Stage 5: Teleoperation Mapping}
\label{subsec:stage5}

\begin{figure} [htbp]
    \centering
    \includegraphics[width=8.7cm]{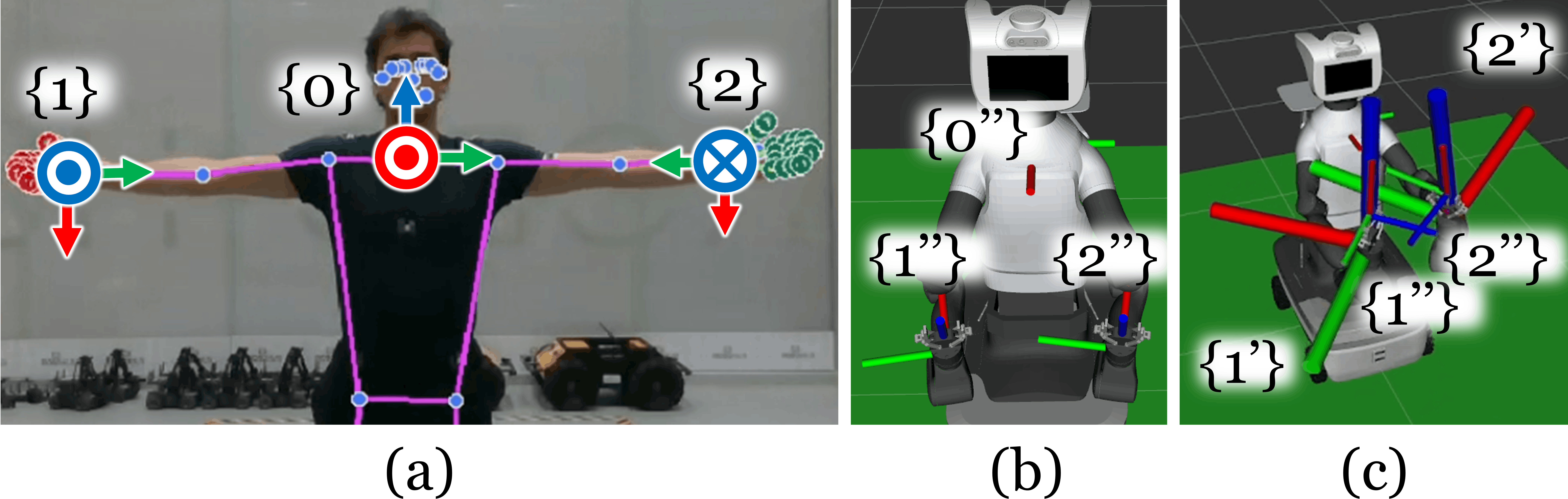}
    \caption{Control frame definitions for teleoperation: (a) operator, (b) robot, and (c) wrist mapping. Frame $\{0\}$ denotes the torso, and frames $\{1\}$ and $\{2\}$ denote the wrists. Frames $\{1'\}$ and $\{2'\}$ represent transformed wrist frames for mapping, while $\{0''\}$, $\{1''\}$, and $\{2''\}$ denote the corresponding robot frames.}
    \label{fig:frame}
\end{figure}

As illustrated in Fig.~\ref{fig:frame}, the AKC-corrected pose is used to construct control frames for motion-mapping teleoperation. A base frame $\{0\}$ is defined at the torso, serving as the global reference for wrist frames $\{1,2\}$. The wrist frame ($f_w$) and torso frame ($f_t$) are defined as:

\vspace{-5mm}
\begin{equation} 
\begin{cases} 
    f_{w,y} = p_e - p_w \\
    f_{w,z} = f_{w,y} \times (p_i - p_p) \\
    f_{w,x} = f_{w,y} \times f_{w,z}
\end{cases},
\begin{cases} 
    f_{t,y} = p_{s,l} - p_{s,r} \\
    f_{t,x} = (p_{h,m} - p_{s,m}) \times f_{t,y}\\
    f_{t,z} = f_{t,x} \times f_{t,y}
\end{cases} 
\end{equation}
\vspace{-5mm}

where $p_{j,m} = (p_{j,l} + p_{j,r})/2$, $j\in \{h,s\}$. $p_i$, $p_p$, and $p_h$ denote positions of index MCP, pinky MCP, and hip.

The wrist frames $\{1,2\}$ are then transformed to  $\{1',2'\}$ for mapping to the robot end-effector frames ( $\{1'',2''\}$) relative to its torso frame $\{0''\}$, as shown in Fig.~\ref{fig:frame}(c). The transformation is defined as a rotation of $-90^\circ$ along the $y$-axis of the wrist frame. For robot motion control, MoveIt with BioIK \cite{bio_ik} is used as the inverse kinematics solver.

The demonstrations of robot teleoperation with Gazebo-simulated and physical robots using the AKC framework are available in the media attachment.

\section{Experimental Validation}
\label{sec:experiment}

In our experiment, ground-truth measurements were acquired using a twelve-camera optical motion capture system (Vicon, Oxford, UK). A participant was instrumented with reflective markers on the upper body following the plug-in gait model \cite{vicon_plugingait}, as shown in Fig.\ref{fig:poses_exp} (a).

Two representative motions involving significant self-occlusion were designed to evaluate the AKC's robustness. Together, these motions enable evaluation under both short- and long-duration occlusion scenarios. Each motion was performed for 30 cycles with a period of 6\,s per cycle. 

\begin{enumerate}[]
    \item \textbf{Motion I (Forward arm movement):} The participant performed a T-pose (Fig.~\ref{fig:poses_exp}(d)), followed by horizontal arm adduction (Fig.~\ref{fig:poses_exp}(e)), and a return to the T-pose. When the arm points toward the RGB-D camera, severe self-occlusion at the wrist, elbow, and shoulder degrades depth measurements.

    \item \textbf{Motion II (Elbow flexion–extension):} The participant performed full forward extension (Fig.~\ref{fig:poses_exp}(f)), followed by elbow flexion to approximately $90^\circ$ (Fig.~\ref{fig:poses_exp}(g)), and returning to full extension. This motion induces prolonged self-occlusion of the shoulder, enabling evaluation under persistent occlusion.
\end{enumerate}

For the AKC configuration, the arm length was initialized in a T-pose, and the average over 100 frames was used to define the preset arm lengths. The first motion cycle was used to determine suitable weights for the evaluation metrics described in Section~\ref{subsec:metric}. To avoid bias, the remaining cycles were used exclusively for performance evaluation in Sections~\ref{subsec:perform_occlusion}, \ref{subsec:filter}, and \ref{subsec:arm length}.

\subsection{System Hardware for Experiment} 

Fig.~\ref{fig:poses_exp} illustrates the experimental setup. The proposed motion capture framework was implemented using an Intel RealSense D435 RGB-D camera operating at a resolution of 480p and a sampling rate of 10~Hz. The system was deployed on a workstation running Ubuntu 22.04, equipped with a 13th-generation Intel Core i9-13900H processor and 32\,GB of RAM. The motion-mapping teleoperation framework was implemented using ROS2 Humble. Experimental validation was conducted using a TIAGo Pro mobile manipulator from PAL Robotics \cite{tiago_robot}.

\subsection{Performance Evaluation under Occlusion} 
\label{subsec:perform_occlusion}

To quantify accuracy, the 3D joint trajectories were evaluated using the root-mean-square error (RMSE) for static motions, and both RMSE and the Pearson correlation coefficient for dynamic motions. Animations of the different mocap results are provided in the attached video (00:25–00:42).

\begin{figure} [htbp]
    \centering
    \includegraphics[width=7cm]{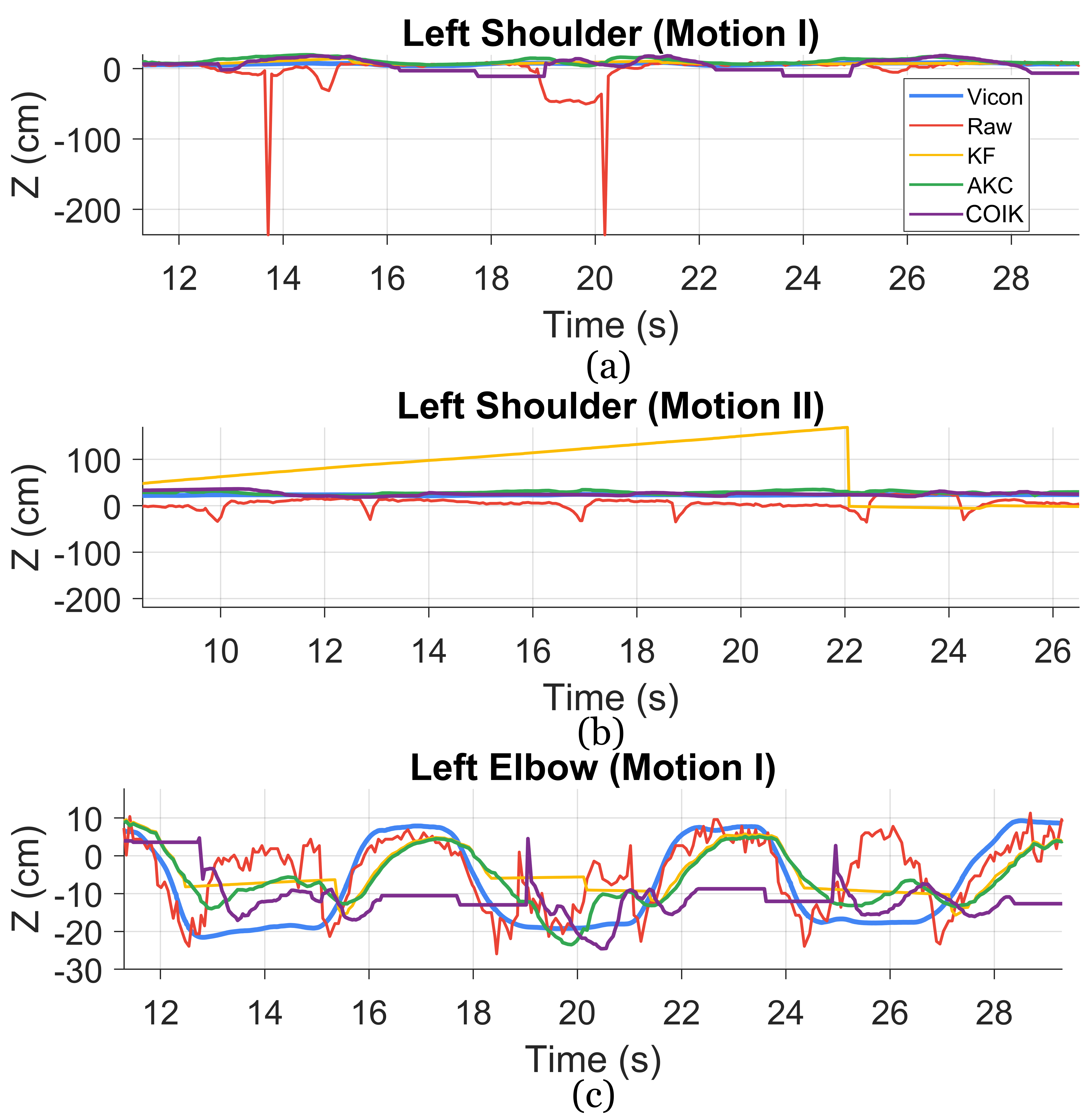}
    \caption{Depth estimation (z) of occluded joints across three example motion cycles. (a-b) Static left shoulder in Motion I and II, and (c) dynamic left elbow in Motion I.}
    \label{fig:occluded}
\end{figure}

Table~\ref{tab:rmse_lsho} presents the average RMSE of an occluded static shoulder under different algorithms. While KF improves depth estimation ($z$) over raw RGB-D measurements in Motion I with short-duration occlusion (Fig.~\ref{fig:occluded}(a)), it does not consistently outperform raw measurements in Motion II under long-duration occlusion. As shown in Fig.~\ref{fig:occluded}(b), the depth estimate exhibits a noticeable linear drift due to the constant-velocity assumption in the KF. Compared to KF, the proposed AKC achieves comparable accuracy in Motion I (within 1.36\,cm) and significantly improved performance in Motion II (over 117\,cm reduction in RMSE). Compared with COIK, AKC provides superior depth estimation while maintaining comparable accuracy in the $x$ and $y$ directions.

\begin{table}[htbp]
\centering
\caption{Avg. RMSE (cm) of an Occluded Static Left Shoulder Compared with Various Algorithms.}
\renewcommand{\arraystretch}{1.2} 
\setlength{\tabcolsep}{5pt}       
\begin{tabular}{l|cccc|cccc}
\hline
& \multicolumn{4}{c|}{\textbf{Motion I}} & \multicolumn{4}{c}{\textbf{Motion II}}\\
\cline{2-9}
 & Raw & \begin{tabular}[c]{@{}c@{}} KF \end{tabular}  & COIK & \begin{tabular}[c]{@{}c@{}} AKC\end{tabular} & Raw & \begin{tabular}[c]{@{}c@{}} KF \end{tabular} & COIK & \begin{tabular}[c]{@{}c@{}} AKC\end{tabular} \\ 
\hline
x & 3.71 & 3.51 & 2.07 & 2.06 & 2.30 & 2.27 & 1.02 & 2.42\\
y & 1.74 & 1.65 & 2.07 & 1.89 & 1.41 & 1.38 & 1.68 & 3.43\\
z & 34.86 & 1.91 & 10.56 & 3.27 & 28.54 & 122.91 & 5.11 & 5.23\\
\hline
\end{tabular}
\label{tab:rmse_lsho}
\end{table}

Table~\ref{tab:rmse_corr_lelb} summarizes the results for the dynamic left elbow in Motion I. Although KF reduces RMSE relative to raw measurements, it does not improve depth correlation. In contrast, AKC achieves similar RMSE to KF but consistently higher correlation, indicating improved temporal consistency. Compared with COIK, AKC yields both lower RMSE and higher correlation, demonstrating more stable and accurate reconstruction under dynamic occlusion. As illustrated in Fig.~\ref{fig:occluded}(c), all methods are affected by noisy raw measurements; however, AKC produces more stable and reliable trajectories than the other approaches.

\begin{table}[htbp]
\centering
\caption{Comparison of Occluded Dynamic Left Elbow in Motion I with Various Algorithms}
\renewcommand{\arraystretch}{1.2}
\setlength{\tabcolsep}{6pt}
\begin{threeparttable}
\begin{tabular}{l|ccccc}
\hline
 & Raw & KF & COIK & AKC \\ 
\hline
Avg. RMSE, x (cm) & 3.06 &  6.16 & 16.88 &  5.33 \\
Avg. RMSE, y (cm) & 5.62 &  5.58 & 5.96 &  5.54  \\
Avg. RMSE, z (cm) & 11.17 &  8.69 & 18.16 &  8.33 \\
\hline
Avg. Corr (x)$^{*}$ & 0.998 &  0.918 & 0.118 &  0.922 \\
Avg. Corr (z)$^{*}$ & 0.888 &  0.774 & -0.113 &  0.832 \\
\hline
\end{tabular}
\begin{tablenotes}
\footnotesize
\item $^{*}$ Pearson correlation coefficient with $p<0.05$.
\end{tablenotes}
\end{threeparttable}
\label{tab:rmse_corr_lelb}
\end{table}

Regarding computational cost, COIK exhibits higher latency and less determinism than AKC. The maximum total computation time across two motions is 491.15\,ms for COIK and 288.84\,ms for AKC. This latency is also evident in the attached video, where delays in COIK affect the motion smoothness. Consequently, COIK may not be suitable for real-time teleoperation applications. In contrast, the additional computational overhead introduced by the AKC layer is minimal (around 11ms), making it more suitable for real-time implementation.

\begin{table}[htbp]
\centering
\caption{Averaged Computation Time (ms) of Various Mocap Algorithms.}
\renewcommand{\arraystretch}{1.2} 
\setlength{\tabcolsep}{6pt}       
\begin{tabular}{l|cccc}
\hline
 & RAW & \begin{tabular}[c]{@{}c@{}} KF \end{tabular}
 & COIK & \begin{tabular}[c]{@{}c@{}} AKC\end{tabular} \\ 
\hline
Motion I & 272.75 & 5.06  & 213.34 & 11.03\\
Motion II & 269.45 & 5.02  & 212.47 & 11.29\\
\hline
\end{tabular}
\label{tab:compute_time}
\end{table}

\subsection{Effect of Temporal Landmark Filtering on AKC} 
\label{subsec:filter}

As discussed in Section~\ref{subsec:stage3}, the AKC method can be integrated with different temporal filtering approaches, such as the KF and EKF. The accuracy of these filters directly affects the AKC performance.

The EKF, which incorporates soft kinematic constraints via a constant-velocity model, provides greater accuracy than the KF. As shown in Fig.~\ref{tab:rmse_rsho_ekf} and Fig.~\ref{tab:rmse_corr_relb_ekf}, the EKF significantly reduces depth errors, achieving a 57\,cm reduction in RMSE for the static shoulder in Motion II. It also improves performance for the dynamic elbow in Motion I, reducing RMSE by 9.8\,cm.

\begin{table}[htbp]
\centering
\caption{Avg. RMSE (cm) of an Occluded Static Right Shoulder Compared with Various Algorithms.}
\renewcommand{\arraystretch}{1.2} 
\setlength{\tabcolsep}{5pt}       
\begin{tabular}{l|cccc|cccc}
\hline
& \multicolumn{4}{c|}{\textbf{Motion I}} & \multicolumn{4}{c}{\textbf{Motion II}} \\
\cline{2-9}
& KF & \makecell{EKF} & \makecell{AKC \\ (KF)} & \makecell{AKC \\ (EKF)} 
& KF & \makecell{EKF} & \makecell{AKC \\ (KF)} & \makecell{AKC \\ (EKF)} \\
\hline
x & 3.48 & 3.49 & 5.56 & 4.14 & 5.38 & 5.47 & 5.71 & 6.04 \\
y & 1.77 & 2.55 & 1.95 & 2.30 & 1.34 & 1.19 & 3.44 & 1.74 \\
z & 2.17 & 2.39 & 6.72 & 2.30 & 65.26 & 8.19 & 9.08 & 8.13 \\
\hline
\end{tabular}
\label{tab:rmse_rsho_ekf}
\end{table}

\begin{table}[htbp]
\centering
\caption{Comparison of Occluded Dynamic Right Elbow in Motion I with Various Algorithms}
\renewcommand{\arraystretch}{1.2}
\setlength{\tabcolsep}{6pt}
\begin{threeparttable}
\begin{tabular}{l|ccccc}
\hline
 & KF & EKF & AKC (KF) & AKC (EKF) \\ 
\hline
Avg. RMSE, x (cm) & 6.25 &  3.82 & 6.13 & 3.80  \\
Avg. RMSE, y (cm) & 3.51 &  2.40 & 3.52 & 2.63  \\
Avg. RMSE, z (cm) & 13.85 &  4.07 & 8.85 & 3.75  \\
\hline
Avg. Corr (x)$^{*}$ & 0.920 &  0.976 & 0.918 &  0.971 \\
Avg. Corr (z)$^{*}$ & 0.772 &  0.977 & 0.850 &  0.978 \\
\hline
\end{tabular}
\begin{tablenotes}
\footnotesize
\item $^{*}$ Pearson correlation coefficient with $p<0.05$.
\end{tablenotes}
\end{threeparttable}
\label{tab:rmse_corr_relb_ekf}
\end{table}

Furthermore, these tables show that AKC, combined with EKF, consistently achieves better motion capture results than when paired with KF, in terms of both RMSE and Pearson correlation. These results indicate that improved temporal filtering enhances the robustness of AKC, particularly in depth estimation under severe occlusion.

When comparing EKF alone with EKF combined with AKC, we observe that AKC further improves depth ($z$) estimation, as measured by both RMSE and Pearson correlation. This indicates that AKC enhances robustness in occlusion. Although slight increases in error are observed in the $x$ and $y$ directions, these changes are minimal (less than 0.65\,cm at worst) and do not significantly affect overall spatial accuracy.

\subsection{AKC Arm Length Variation} 
\label{subsec:arm length}

To capture realistic human motion, maintaining consistent arm length is essential for anatomical plausibility. Therefore, we compare the AKC with existing approaches. As shown in Table~\ref{tab:arm_length_range}, AKC maintains constant arm length throughout the motion. In contrast, the COIK may fail to preserve consistent arm length when the optimization reaches its iteration limit. Raw depth measurements, as well as KF and EKF, do not enforce this constraint and thus exhibit noticeable variations in arm length.

\begin{table}[htbp]
\centering
\caption{Range of Arm Length Variation During Motion I \& II (cm)}
\renewcommand{\arraystretch}{1.2}
\setlength{\tabcolsep}{6pt}
\resizebox{\columnwidth}{!}{
\begin{tabular}{l|ccccccc}
\hline
 & Vicon & Raw & KF & EKF & COIK I$^{*}$ & COIK II$^{+}$ & AKC$^{\#}$ \\ 
\hline
LUA & 4.0 & 242.4 & 507.4 & 25.3 & 1.4 & 0.0 & 0.0 \\
LLA & 2.0 & 222.5 & 34.3 & 18.6 & 2.3 & 0.0 & 0.0 \\
RUA & 3.7 & 232.9 & 407.0 & 58.9 & 2.6 & 0.0 & 0.0 \\
RLA & 2.2 & 194.7 & 31.2 & 16.5 & 3.9 & 0.0 & 0.0 \\
\hline
\end{tabular}
}
\begin{tablenotes}
\footnotesize
\item $^{*}$ COIK when iteration limit reached (Mostly Motion I)
\item $^{+}$ COIK when feasible solutions can be obtained
\item $^{\#}$ Both AKC with KF and EKF
\end{tablenotes}

\label{tab:arm_length_range}
\end{table}

Notably, even the Vicon system, despite using multiple cameras, shows a 4cm variation in measured arm length due to marker placement errors and soft-tissue motion artifacts.

\subsection{AKC Metric Weight Sensitivity Analysis} 
\label{subsec:metric_effect}

This section uses the first cycle of Motion I and Motion II to analyze the metric effect on candidate selection and to determine suitable weights. As described in Section~\ref{subsec:metric}, five metrics are used to select the optimal configuration from four elbow–shoulder candidate pairs.

We focus on three primary metrics that directly influence accuracy: (i) elbow spatial consistency, (ii) shoulder spatial consistency, and (iii) temporal consistency. The remaining two metrics primarily eliminate anatomically infeasible candidates and therefore can be assigned relatively large weights (i.e., $w_4 = w_5 =100.0$).

\begin{figure} [htbp]
    \centering
    \includegraphics[width=8.7cm, height=9.8cm]{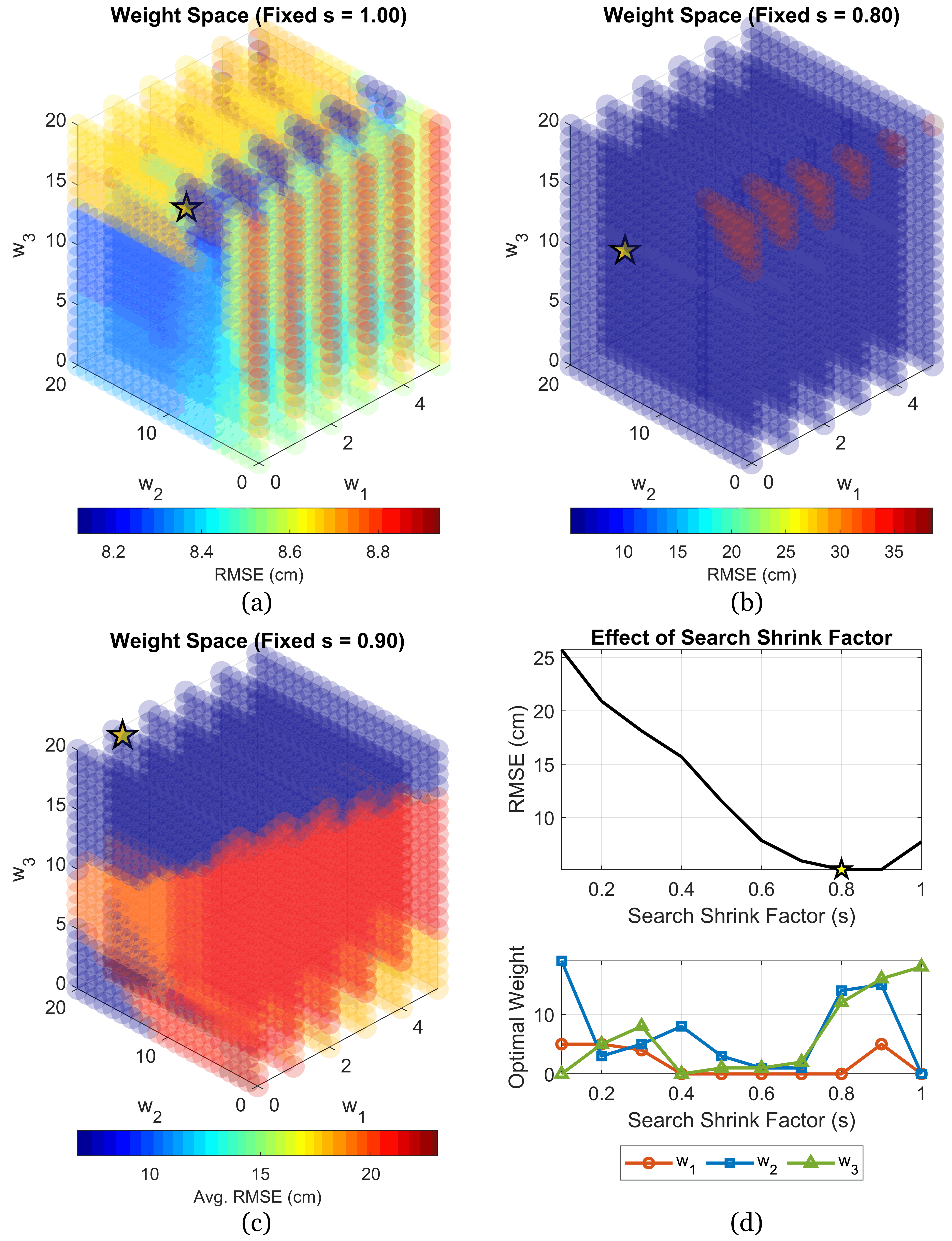}
    \caption{Effect of three metrics (i.e., spatial consistency of shoulder and elbow candidates, and temporal consistency) on the averaged L2 norm of 3D joint position error of both arms during (a) Motion I, (b) Motion II, and (c) both Motion I and Motion II. (d) Effect of search shrink factor during Motion II.}
    \label{fig:metrics}
\end{figure}

The metric weights are found to be motion-dependent, as illustrated in Fig.~\ref{fig:metrics}. By minimizing the L2-norm joint error, optimal weights were obtained: 

\begin{enumerate}[]
    \item \textbf{Motion I ($W_a$):} $w_1 = 0.0$, $w_2 = 8.0$, $w_3 = 18.0$, $s = 1.0$; 
    \item \textbf{Motion II ($W_b$):} $w_1 = 0.0$, $w_2 = 14.0$, $w_3 = 12.0$, $s = 0.8$;  
    \item \textbf{Motion I \& II ($W_c$):} $w_1 = 1.0$, $w_2 = 19.0$, $w_3 = 20.0$, $s = 0.9$.
\end{enumerate}

Overall, the results show that the metric weights $(w_1, w_2, w_3)$ are relatively insensitive within a reasonable range, producing comparable results for a fixed search shrink factor ($s$). In contrast, the $s$ has a more significant impact on performance. As shown in Fig.~\ref{fig:metrics}(d), suboptimal values of $s$ can lead to large errors, with RMSE increasing up to 25\,cm.

To further evaluate robustness, weights calibrated on one motion were applied to the other motion (unseen scenario). The results in Table~\ref{tab:metric_lsho_lelb} show that larger $s$ (e.g., $s = 1.0$ in $W_a$) are less robust under unseen conditions (Motion II), leading to increased RMSE. In contrast, a moderate value ($s = 0.8$ in $W_b$) provides more consistent performance across both seen (Motion II) and unseen motions (Motion I), compared to $s = 1.0$ in $W_a$ and $s = 0.9$ in $W_c$. However, excessively small values of $s$ (i.e., $0.1$ - $0.5$) degrade performance, as illustrated in Fig.~\ref{fig:metrics}(d) and the attached video (00:43–00:58). Based on these observations, we adopt $s = 0.8$ as a robust choice.

\begin{table}[htbp]
\centering
\caption{RMSE (cm) of 29 Motion Cycles with Different Optimal Weights ($W_a$,$W_b$,$W_c$) for Left Arm.}
\renewcommand{\arraystretch}{1.2} 
\setlength{\tabcolsep}{9pt}       
\begin{tabular}{l|cc|cc}
\hline
& \multicolumn{2}{c|}{\textbf{Motion I using $W_a$}} & \multicolumn{2}{c}{\textbf{Motion II using $W_b$}} \\ 
\cline{2-5}
 & Shoulder & \begin{tabular}[c]{@{}c@{}} Elbow \end{tabular} 
 & Shoulder & \begin{tabular}[c]{@{}c@{}} Elbow\end{tabular} \\ 
\hline
x & 3.60$\pm$0.24 & 6.20$\pm$0.44 & 2.42$\pm$0.85 & 1.53$\pm$0.67 \\
y & 1.25$\pm$0.81 & 5.27$\pm$0.89 & 3.43$\pm$1.47 & 3.01$\pm$1.29 \\
z & 9.37$\pm$11.87 & 7.72$\pm$0.74 & 5.23$\pm$2.09 & 4.98$\pm$2.00 \\
\hline

& \multicolumn{2}{c|}{\textbf{Motion I using $W_b$}} & \multicolumn{2}{c}{\textbf{Motion II using $W_a$}} \\ 
\cline{2-5}
 & Shoulder & \begin{tabular}[c]{@{}c@{}} Elbow \end{tabular} 
 & Shoulder & \begin{tabular}[c]{@{}c@{}} Elbow\end{tabular} \\ 
\hline
x & 2.06$\pm$0.34 & 5.33$\pm$0.41 & 1.79$\pm$0.52 & 1.45$\pm$0.61 \\
y & 1.89$\pm$0.74 & 5.54$\pm$1.30 & 2.90$\pm$1.66 & 2.95$\pm$1.13 \\
z & 3.27$\pm$1.06 & 8.33$\pm$0.77 & 23.05$\pm$16.21 & 18.35$\pm$13.54 \\
\hline

& \multicolumn{2}{c|}{\textbf{Motion I using $W_c$}} & \multicolumn{2}{c}{\textbf{Motion II using $W_c$}} \\ 
\cline{2-5}
 & Shoulder & \begin{tabular}[c]{@{}c@{}} Elbow \end{tabular} 
 & Shoulder & \begin{tabular}[c]{@{}c@{}} Elbow\end{tabular} \\ 
\hline

x & 2.05$\pm$0.30 & 5.24$\pm$0.50 & 2.04$\pm$0.60 & 1.65$\pm$0.93 \\
y & 1.29$\pm$0.69 & 4.97$\pm$0.73 & 2.80$\pm$1.51 & 2.52$\pm$1.06 \\
z & 4.17$\pm$1.50 & 9.06$\pm$1.85 & 8.56$\pm$9.15 & 8.55$\pm$9.50 \\
\hline

\end{tabular}
\label{tab:metric_lsho_lelb}
\end{table}

\subsection{Effect of Arm-Length Calibration Methods} 
\label{sec:arm_length_effect}

This section evaluates four calibration approaches for determining the preset arm length and their impact on performance.

\begin{enumerate}[]
    \item \textbf{Manual} — Arm length measured with a ruler before the experiment. This approach is simple but prone to measurement inaccuracies.
    \item \textbf{Vicon} — Arm length estimated by Plug-in Gait Dynamic pipeline of the Vicon system. 
    \item \textbf{T-pose} — Arm length computed from KF-filtered landmark positions obtained with our mocap framework while the participant held a T-pose for 100 frames (Fig.~\ref{fig:poses_exp}(b)).  
    \item \textbf{L-pose} — Similar to the T-pose method, but with the participant holding an L-pose (Fig.~\ref{fig:poses_exp}(c)).
\end{enumerate}

As shown in Table~\ref{tab:arm_length}, the four calibration methods yield different arm length estimates, and Table~\ref{tab:rmse_arm_length} summarizes the corresponding RMSE between reconstructed joint trajectories and the Vicon reference. The results indicate that the average RMSE is comparable across different calibration approaches.

\begin{table}[htbp]
\centering
\caption{Arm Length for 4 Arm-Length Measurement Methods (cm)}
\renewcommand{\arraystretch}{1.2}
\setlength{\tabcolsep}{6pt}
\begin{threeparttable}
\begin{tabular}{l|ccccc}
\hline
 & Manual & Vicon & T-pose & L-pose \\ 
\hline
LUA$^{1}$ & 32.0 & 28.8 & 28.5 & 28.4 \\
RUA$^{3}$ & 32.0 & 28.9 & 29.5 & 29.9 \\
LLA$^{2}$ & 26.0 & 28.0 & 27.0 & 26.4 \\
RLA$^{4}$ & 26.0 & 27.0 & 26.9 & 27.5 \\
SW$^{5}$ & 33.0 & 33.2 & 36.6 & 36.7 \\
\hline
\end{tabular}
\begin{tablenotes}[flushleft]
\footnotesize
\item $L$:Left, $R$:Right, $UA$:Upper Arm, $LA$:Lower Arm, $^{5}$ Shoulder Width.
\end{tablenotes}
\end{threeparttable}
\label{tab:arm_length}
\end{table}

\begin{table}[htbp]
\centering
\caption{Averaged RMS Euclidean Distance Errors (cm) and L2-normalized errors for 4 Measurement Methods}
\renewcommand{\arraystretch}{1.2}
\setlength{\tabcolsep}{6pt}
\begin{threeparttable}
\begin{tabular}{l|ccccc}
\hline
 & Manual & Vicon & T-pose & L-pose \\
\hline
$LELB$ & 11.6 & 10.7 & 10.7 & 12.7 \\
$RELB$ & 9.8 & 8.5  & 10.3 & 9.5 \\
$LSHO$ & 3.5 & 4.0  & 4.0 & 6.3 \\
$RSHO$ & 7.8 & 6.9  & 8.1 & 7.5 \\
\hline
L2-norm & 17.4 & 15.8 & 17.4 & 18.6 \\
\hline
\end{tabular}
\begin{tablenotes}[flushleft]
\footnotesize
\item $L$:Left, $R$:Right, $ELB$:Elbow, $SHO$:Shoulder.
\end{tablenotes}
\end{threeparttable}
\label{tab:rmse_arm_length}
\end{table}

From a practical perspective, methods relying on external systems (e.g., Vicon) or manual measurements are less suitable for real-world deployment. In contrast, calibration using the same RGB-D system (i.e., T-pose and L-pose) offers a more convenient, self-contained solution for future applications.

\subsection{Motion Mapping for Robot Teleoperation}
\label{subsec:teleop}

We demonstrate the integration of AKC (with KF) into the robot teleoperation framework. As described in Section~\ref{subsec:stage5}, the AKC-corrected pose is used to construct robot control frames. The evaluation consists of three parts: (i) control frame construction with different mocap algorithms, (ii) simulated teleoperation in Gazebo, and (iii) real robot teleoperation for a pick-and-place task.

The attached video (01:00–01:16) compares wrist frames relative to the torso using different mocap methods, including raw RGB-D measurements, KF, AKC with KF, and COIK. The results show that AKC with KF provides the most stable and consistent control frames. In contrast, raw measurements and KF are significantly affected by self-occlusion and inconsistent arm lengths, leading to large variations in the constructed frames that are unsuitable for teleoperation.

Despite its ability to generate many candidate solutions, COIK exhibits unstable behavior. When the optimization converges to suboptimal local minima, the resulting configurations are unreliable. Moreover, the weighting parameter $\lambda$ in Eqn.~\ref{eqn:coik} is more sensitive to tuning compared to AKC. Combined with the higher computational latency reported in Table~\ref{tab:compute_time}, these factors make COIK less suitable for real-time teleoperation applications.

\begin{figure} [htbp]
    \centering
    \includegraphics[width=7cm]{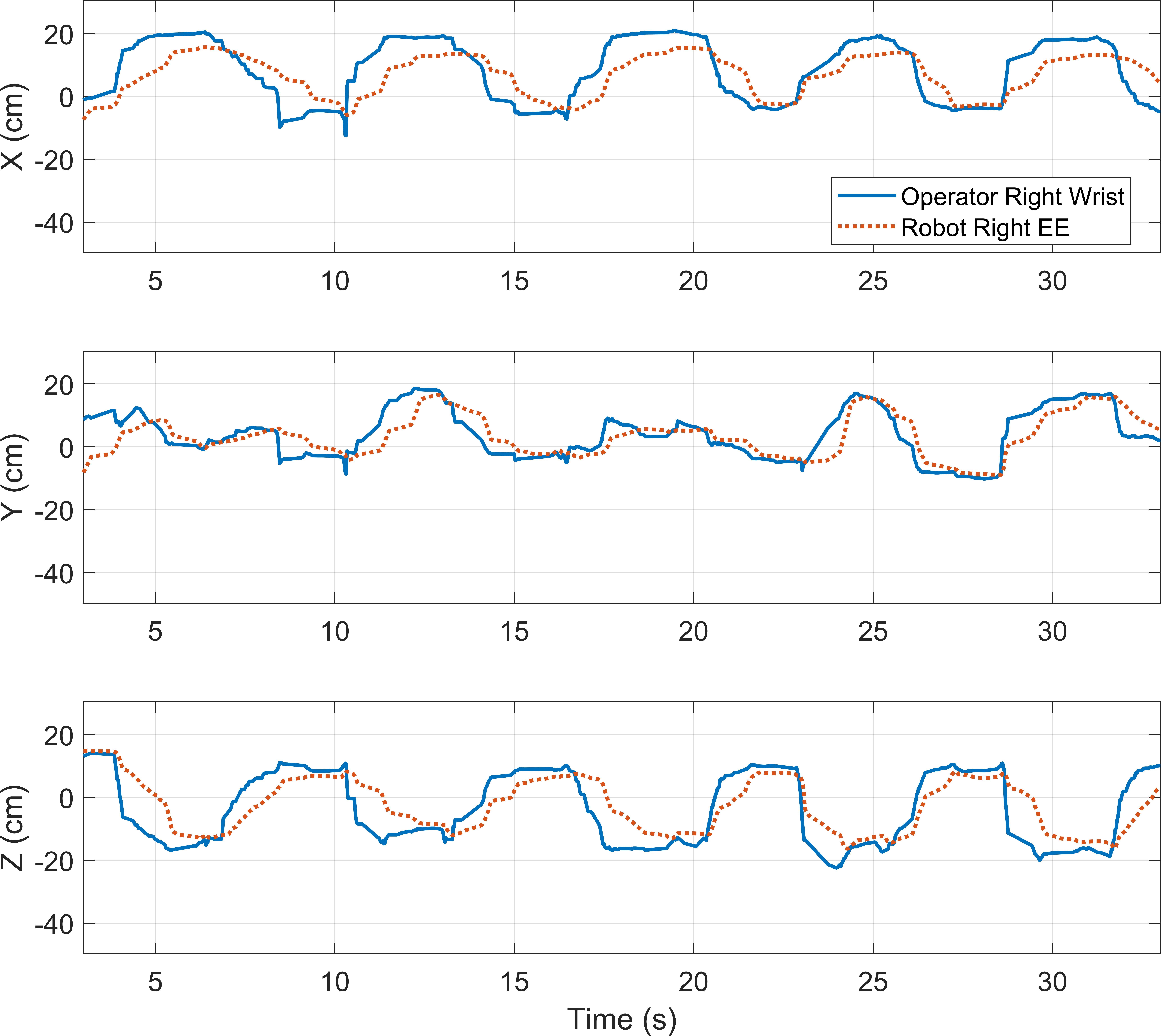}
    \caption{Motion of the right robot end-effector in simulated teleoperation, controlled by the operator’s right wrist.}
    \label{fig:teleop}
\end{figure}

For teleoperation in Gazebo, the attached video (01:17–01:30) and Fig.~\ref{fig:teleop} demonstrate that an operator can successfully control a simulated robot using AKC with KF, even under shoulder and elbow occlusion. 

Furthermore, the video (01:31–02:07) presents real-time teleoperation, in which the operator controls a physical robot to perform a pick-and-place task with an irregularly shaped object. During the task, the operator intentionally introduced elbow occlusion by randomly moving the arm before grasping the object, then raised the arm to induce shoulder occlusion. 

Despite these challenging conditions, the system maintained stable control using AKC with KF. However, minor perturbations are observed due to the limitations of the KF-based temporal filtering. As discussed in Section~\ref{subsec:filter}, further improvements in motion capture accuracy can be achieved by employing more advanced filtering methods, such as the EKF.

\section{Discussion}
\label{sec:discussion}

Conventional motion capture systems are often complex and intrusive, requiring multiple cameras and markers. In contrast, single RGB-D camera systems offer a low-cost, non-invasive alternative that is more suitable for deployment in constrained or real-world environments. However, their performance is significantly degraded by self-occlusion, particularly during upper-limb motion, resulting in inaccurate depth estimates and anatomically inconsistent poses.

The AKC addresses this challenge by enforcing geometric constraints based on constant arm lengths. By leveraging the more reliable wrist position and reconstructing occluded joints using the Pythagorean theorem, AKC improves depth estimation while preserving the accuracy of image-plane coordinates. Experimental results show that AKC significantly reduces depth errors while maintaining arm-length consistency, without introducing substantial distortion in the $x$ and $y$ directions. This capability is critical for downstream applications such as teleoperation, where depth inconsistencies can lead to distorted robot control frames and inverse-kinematics instability.

A key advantage of AKC is its deterministic formulation. Unlike optimization-based approaches, which rely on iterative solvers and sensitive parameter tuning, AKC generates a small number of candidate solutions and selects the most plausible configuration using lightweight evaluation metrics. This reduces the solution space from infinitely many possibilities to four candidates, significantly lowering computational cost and enabling predictable behavior. As a result, AKC is suitable for real-time applications.

Compared with existing methods, AKC offers a simpler and more robust alternative for enforcing anatomical constraints. Prior work \cite{LEE1985148} utilized constant arm-length assumptions to construct feasible configurations, but relied on complex interpretation trees and restrictive assumptions such as constant shoulder width, which may not hold due to scapular motion. Optimization-based approaches \cite{kim_human_2023} incorporate arm-length constraints but are computationally expensive and sensitive to weight tuning, and may converge to suboptimal solutions. In contrast, AKC directly enforces geometric consistency through a closed-form formulation, achieving stable and efficient reconstruction even under severe occlusion.

AKC functions as a geometric correction layer that can be integrated with different temporal filtering methods. While filters such as KF and EKF improve temporal smoothness, they may still produce inconsistent depth estimates due to noise or model limitations. AKC addresses this limitation by enforcing geometric consistency, thereby correcting residual depth errors that filtering alone cannot resolve. Its performance further benefits from improved filtering accuracy, as more reliable inputs lead to better candidate selection. This design allows the framework to be flexibly adapted to different filtering methods and applications without modifying the AKC algorithm.

Another practical advantage of AKC is its minimal setup requirement. The method requires only a one-time calibration of arm length, which can be performed using the same RGB-D camera. This is simpler than multi-camera marker-based systems and enables easier deployment in real-world scenarios.

The proposed AKC method can be extended to lower-limb motion capture, although this work focuses on upper-limb motion-mapping teleoperation. The current formulation assumes that wrist positions are relatively reliable, as they are typically closer to the camera and less prone to occlusion, and thus uses the wrist as the reference point for reconstruction. Extending this approach to the lower limb requires selecting an appropriate reference joint. However, this is more challenging due to greater variability in motion and occlusion patterns. For example, the knee may occlude the hip in seated postures, while the ankle may be occluded during walking. As a result, a single reference joint may not be consistently reliable across different motions. Addressing these challenges requires further investigation and is left for future work, as it is beyond the scope of this study.

Beyond teleoperation, the proposed framework has important implications for assistive robotics and remote caregiving. While our primary motivation is to enable caregivers to remotely control robots, this system can also empower older adults or individuals with limited mobility. For example, users who cannot physically access stores may use motion-mapping teleoperation to control a robot to perform tasks such as grocery retrieval \cite{Tomizawa_2007}. The minimal setup requirement and robustness under occlusion make the proposed approach suitable for such assistive applications.

In summary, the AKC provides a computationally efficient, geometrically grounded, and easily deployable solution for improving RGB-D-based motion capture under occlusion. Its deterministic nature, compatibility with filtering methods, and reliable performance in real-time settings make it a practical approach for human–robot interaction and teleoperation in real-world environments.

\section{Conclusion}
\label{sec:conclusion}

This paper presents the proposed Arm Kinematic Correction (AKC) method as a post-processing framework for single RGB-D camera-based motion capture. The method improves depth estimation for occluded joints by enforcing geometric constraints based on constant arm lengths. Unlike optimization-based approaches, AKC adopts a deterministic formulation that avoids complex probabilistic modeling and extensive parameter tuning. 

Experimental results demonstrate that the proposed method enhances robustness and maintains anatomical consistency, particularly under severe self-occlusion. Furthermore, integrating AKC into a motion-mapping teleoperation system demonstrates its effectiveness in enabling stable, reliable robot control. These properties make AKC a practical and efficient solution for applications such as robot teleoperation in constrained or occlusion-prone environments.

\bibliographystyle{IEEEtran}
\bibliography{references}

@misc{noauthor_ageing_nodate,
  author = "{World Health Organization}",
  title = "{Ageing and health}",
  year = {2024},
  url = {https://www.who.int/news-room/fact-sheets/detail/ageing-and-health},
  note = "Accessed: Feb. 7, 2025"
}

@misc{prb_ageing,
  author = "{Population Reference Bureau}",
  title = "{2024 World Population Data Sheet.}",
  year = {2024},
  url = {https://2024-wpds.prb.org/wp-content/uploads/2024/09/2024-World-Population-Data-Sheet-Booklet.pdf},
  note = "Accessed: Mar. 25, 2025"
}

@article{lyu_teleoperation_2020,
	title = {Teleoperation of {Collaborative} {Robot} for {Remote} {Dementia} {Care} in {Home} {Environments}},
	volume = {8},
	issn = {2168-2372},
	doi = {10.1109/JTEHM.2020.3002384},
	journal = {IEEE Journal of Translational Engineering in Health and Medicine},
	author = {Lyu et al., Honghao},
	year = {2020},
	pages = {1--10},
}

@inproceedings{shah_toward_2024,
	title = {Toward {Intelligent} {Telepresence} {Robotics} for {Enhancing} {Elderly} {Healthcare} in {Smart} {Care} {Home}},
	isbn = {978-3-031-60412-6},
	doi = {10.1007/978-3-031-60412-6_14},
	language = {en},
	booktitle = {Human-{Computer} {Interaction}},
	publisher = {Springer Nature},
	author = {Shah et al., Syed Hammad Hussain},
	year = {2024},
	pages = {180--195},
}

@article{bazarevsky_blazepose_2020,
	title = {{BlazePose}: {On}-device {Real}-time {Body} {Pose} tracking},
	doi = {10.48550/arXiv.2006.10204},
	journal = {arXiv},
	author = {Bazarevsky et al., Valentin},
	month = jun,
	year = {2020}
}

@misc{kwok_leveraging_2025,
	title = {Leveraging {GCN}-based {Action} {Recognition} for {Teleoperation} in {Daily} {Activity} {Assistance}},
	doi = {10.48550/arXiv.2504.07001},
	publisher = {arXiv},
	author = {Kwok et al., Thomas M.},
	month = apr,
	year = {2025},
}

@article{zhou_review_2024,
	title = {A {Review} of {Depth}-{Based} {Human} {Motion} {Enhancement}: {Past} and {Present}},
	volume = {28},
	issn = {2168-2208},
	doi = {10.1109/JBHI.2023.3257662},
	number = {2},
	journal = {IEEE Journal of Biomedical and Health Informatics},
	author = {Zhou et al, Le},
	month = feb,
	year = {2024},
	pages = {633--644},
}

@article{kim_human_2023,
	title = {Human {Pose} {Estimation} {Using} {MediaPipe} {Pose} and {Optimization} {Method} {Based} on a {Humanoid} {Model}},
	volume = {13},
	copyright = {http://creativecommons.org/licenses/by/3.0/},
	issn = {2076-3417},
	url = {https://www.mdpi.com/2076-3417/13/4/2700},
	doi = {10.3390/app13042700},
	number = {4},
	journal = {Applied Sciences},
	author = {Kim et al., Jong-Wook},
	month = jan,
	year = {2023},
	pages = {2700},
}

@INPROCEEDINGS{kf_paper,
  author={Buizza et al., Caterina},
  booktitle={2020 IEEE Winter Conference on Applications of Computer Vision (WACV)}, 
  title={Real-Time Multi-Person Pose Tracking using Data Assimilation}, 
  year={2020},
  volume={},
  number={},
  pages={438-447},
  doi={10.1109/WACV45572.2020.9093442}}

@misc{vicon_plugingait,
  title        = {Vicon Plug-in Gait Reference Guide},
  author = {{Vicon Motion Systems Ltd.}},
  url = {https://help.vicon.com/space/Nexus216/11607059},
  note         = {Accessed: July 28, 2025},
  year         = {2025}
}

@misc{mediapipe_pose_landmarker,
  author       = {{Google AI}},
  title        = {MediaPipe Pose Landmarker for Python},
  year         = {2025},
  howpublished = {\url{https://ai.google.dev/edge/mediapipe/solutions/vision/pose_landmarker/python}},
  note         = {Accessed: 28 July 2025}
}

@article{bin_survey_2025,
	title = {A survey on the visual perception of humanoid robot},
	volume = {5},
	issn = {2667-3797},
	doi = {10.1016/j.birob.2024.100197},
	number = {1},
	urldate = {2025-07-29},
	journal = {Biomimetic Intelligence and Robotics},
	author = {Bin et al., Teng},
	month = mar,
	year = {2025},
	pages = {100197},
}

@article{lannan_human_2022,
	title = {Human {Motion} {Enhancement} via {Tobit} {Kalman} {Filter}-{Assisted} {Autoencoder}},
	volume = {10},
	issn = {2169-3536},
	doi = {10.1109/ACCESS.2022.3157605},
	urldate = {2025-07-29},
	journal = {IEEE Access},
	author = {Lannan et al., Nate},
	year = {2022},
	pages = {29233--29251},
}

@inproceedings{tripathy_constrained_2018,
	title = {Constrained {Particle} {Filter} for {Improving} {Kinect} {Based} {Measurements}},
	doi = {10.23919/EUSIPCO.2018.8553437},
	booktitle = {2018 26th {European} {Signal} {Processing} {Conference} ({EUSIPCO})},
	author = {Tripathy et al., Soumya Ranjan},
	month = sep,
	year = {2018},
	note = {ISSN: 2076-1465},
	pages = {306--310},
}

@article{moon_multiple_2016,
	title = {Multiple {Kinect} {Sensor} {Fusion} for {Human} {Skeleton} {Tracking} {Using} {Kalman} {Filtering}},
	volume = {13},
	issn = {1729-8806},
	doi = {10.5772/62415},
	language = {EN},
	number = {2},
	journal = {International Journal of Advanced Robotic Systems},
	author = {Moon et al., Sungphill},
	month = mar,
	year = {2016},
	note = {Publisher: SAGE Publications},
	pages = {65},
}

@inproceedings{das_improvement_2017,
	title = {Improvement in {Kinect} based measurements using anthropometric constraints for rehabilitation},
	doi = {10.1109/ICC.2017.7996969},
	booktitle = {2017 {IEEE} {International} {Conference} on {Communications} ({ICC})},
	author = {Das et al., Pratyusha},
	month = may,
	year = {2017},
	note = {ISSN: 1938-1883},
	pages = {1--6},
}

@article{wang_spatio_temporal_2021,
	title = {Spatio-{Temporal} {Manifold} {Learning} for {Human} {Motions} via {Long}-{Horizon} {Modeling}},
	volume = {27},
	issn = {1077-2626},
	doi = {10.1109/TVCG.2019.2936810},
	number = {1},
	journal = {IEEE Transactions on Visualization and Computer Graphics},
	author = {Wang et al., He},
	month = jan,
	year = {2021},
	pages = {216--227},
}

@misc{realsense_depth_guide,
  author       = {{Intel RealSense}},
  title        = {Beginner's Guide to Depth},
  howpublished = {[Online]. Available: \url{https://www.realsenseai.com/stereo-depth/beginners-guide-to-depth/}},
  note         = {Accessed: May 26, 2026}
}

@article{LEE1985148,
title = {Determination of 3D human body postures from a single view},
journal = {Computer Vision, Graphics, and Image Processing},
volume = {30},
number = {2},
pages = {148-168},
year = {1985},
issn = {0734-189X},
doi = {https://doi.org/10.1016/0734-189X(85)90094-5},
url = {https://www.sciencedirect.com/science/article/pii/0734189X85900945},
author = {Hsi-Jian Lee and Zen Chen},
}

@misc{mediapipe_holistic,
  author       = {{Google AI}},
  title        = {MediaPipe Holistic},
  howpublished = {\url{https://github.com/google-ai-edge/mediapipe/blob/master/docs/solutions/holistic.md}},
  note         = {Accessed: 28 July 2025}
}

@INPROCEEDINGS{bio_ik,
  author={Ruppel, Philipp and Hendrich, Norman and Starke, Sebastian and Zhang, Jianwei},
  booktitle={2018 IEEE International Conference on Robotics and Automation (ICRA)}, 
  title={Cost Functions to Specify Full-Body Motion and Multi-Goal Manipulation Tasks}, 
  year={2018},
  pages={3152-3159},
  doi={10.1109/ICRA.2018.8460799}
}

@misc{tiago_robot,
  author       = {{PAL Robotics}},
  title        = {TIAGo Pro - Empowering Mobile Manipulation},
  howpublished = {\url{https://pal-robotics.com/robot/tiago-pro/}},
  note         = {Accessed: 3 Jun 2026}
}

@INPROCEEDINGS{Tomizawa_2007,
  author={Tomizawa, Tetsuo and Ohba, Kohtaro and Ohya, Akihisa and Yuta, Shin'ichi},
  booktitle={2007 International Conference on Mechatronics and Automation}, 
  title={Remote Food Shopping Robot System in a Supermarket -Realization of the shopping task from remote places}, 
  year={2007},
  pages={1771-1776},
  doi={10.1109/ICMA.2007.4303818}}

@article{chromy_2025,
  title={Validated low-cost standardized VICON configuration as a practical approach to estimating the minimal accuracy of a specific setup},
  author={Chromy, Adam and Sopak, Petr and Cigler, Hynek},
  journal={Scientific Reports},
  volume={15},
  number={1},
  pages={23351},
  year={2025},
  publisher={Nature Publishing Group UK London}
}

@article{kakavand_2025,
  title={Comparison of kinematics and kinetics between OpenCap and a marker-based motion capture system in cycling},
  author={Kakavand, Reza and Ahmadi, Reza and Parsaei, Atousa and Edwards, W Brent and Komeili, Amin},
  journal={Computers in Biology and Medicine},
  volume={192},
  pages={110295},
  year={2025},
  publisher={Elsevier}
}

@article{federolf_2025,
  title={Validation of markerless (Theia3D™) against marker-based (Vicon™) motion capture data of postural control movements analyzed through principal component analysis},
  author={Federolf, Peter and K{\"u}hne, Mareike and Schiel, Katharina and Reimeir, Elisa and Debertin, Daniel and Calisti, Mait{\'e} and Mohr, Maurice},
  journal={Journal of Biomechanics},
  volume={189},
  pages={112831},
  year={2025},
  publisher={Elsevier}
}

@article{pham_2025,
  title={3D Human Pose Estimation Using Body Kinematics and Kalman Filter},
  author={Pham, Thanh Dat and David, Deepak Antony and Busse, Luke and Omotuyi, Oyindamola and Kumar, Manish},
  journal={IFAC-PapersOnLine},
  volume={59},
  number={30},
  pages={156--161},
  year={2025},
  publisher={Elsevier}
}

@article{kalman_1960,
  title={A new approach to linear filtering and prediction problems},
  author={Kalman, Rudolph Emil},
  year={1960}
}

@article{welch_1995,
  title={An introduction to the Kalman filter},
  author={Welch, Greg and Bishop, Gary and others},
  year={1995},
  publisher={Chapel Hill, NC, USA}
}

@inproceedings{joukov_2018,
  title={Real-time unlabeled marker pose estimation via constrained extended Kalman filter},
  author={Joukov, Vladimir and Lin, Jonathan FS and Westermann, Kevin and Kuli{\'c}, Dana},
  booktitle={International Symposium on Experimental Robotics},
  pages={762--771},
  year={2018},
  organization={Springer}
}

@inproceedings{kwok_2023,
  title={A reliable kinematic measurement of upper limb exoskeleton for VR therapy with visual-inertial sensors},
  author={Kwok, Thomas M and Li, Tong and Yu, Haoyong},
  booktitle={2023 IEEE/ASME International Conference on Advanced Intelligent Mechatronics (AIM)},
  pages={584--590},
  year={2023},
  organization={IEEE}
}

@article{tang_1999,
  title={A constrained inverse kinematics technique for real-time motion capture animation},
  author={Tang, Wen and Cavazza, Marc and Mountain, Dale and Earnshaw, Rae},
  journal={The Visual Computer},
  volume={15},
  number={7},
  pages={413--425},
  year={1999},
  publisher={Springer}
}

@inproceedings{cotton_2023,
  title={Optimizing trajectories and inverse kinematics for biomechanical analysis of markerless motion capture data},
  author={Cotton, R James and DeLillo, Allison and Cimorelli, Anthony and Shah, Kunal and Peiffer, JD and Anarwala, Shawana and Abdou, Kayan and Karakostas, Tasos},
  booktitle={2023 International Conference on Rehabilitation Robotics (ICORR)},
  pages={1--6},
  year={2023},
  organization={IEEE}
}

@inproceedings{petrich_2022,
  title={A quantitative analysis of activities of daily living: Insights into improving functional independence with assistive robotics},
  author={Petrich, Laura and Jin, Jun and Dehghan, Masood and Jagersand, Martin},
  booktitle={2022 International Conference on Robotics and Automation (ICRA)},
  pages={6999--7006},
  year={2022},
  organization={IEEE}
}

@ARTICLE{Castro_2026,
  author={Castro, John W. and Ariza, Víctor and Riffo, Vladimir},
  journal={IEEE Access}, 
  title={Understanding Teleoperation in Robotics: A Systematic Review of Device Types and Main Performance}, 
  year={2026},
  volume={14},
  number={},
  pages={45377-45409},
  doi={10.1109/ACCESS.2026.3676696}}

@article{openpose_2019,
  author = {Z. {Cao} and G. {Hidalgo Martinez} and T. {Simon} and S. {Wei} and Y. A. {Sheikh}},
  journal = {IEEE Transactions on Pattern Analysis and Machine Intelligence},
  title = {OpenPose: Realtime Multi-Person 2D Pose Estimation using Part Affinity Fields},
  year = {2019}
}

@article{vskulj2021wearable,
  title={A wearable imu system for flexible teleoperation of a collaborative industrial robot},
  author={{\v{S}}kulj, Ga{\v{s}}per and Vrabi{\v{c}}, Rok and Podr{\v{z}}aj, Primo{\v{z}}},
  journal={Sensors},
  volume={21},
  number={17},
  pages={5871},
  year={2021},
  publisher={MDPI}
}

@Article{robotics12060163,
AUTHOR = {Galarza, Bryan R. and Ayala, Paulina and Manzano, Santiago and Garcia, Marcelo V.},
TITLE = {Virtual Reality Teleoperation System for Mobile Robot Manipulation},
JOURNAL = {Robotics},
VOLUME = {12},
YEAR = {2023},
NUMBER = {6},
ARTICLE-NUMBER = {163},
URL = {https://www.mdpi.com/2218-6581/12/6/163},
ISSN = {2218-6581},
DOI = {10.3390/robotics12060163}
}

@ARTICLE{9133071,
  author={Chattha, Umer Asghar and Janjua, Uzair Iqbal and Anwar, Fozia and Madni, Tahir Mustafa and Cheema, Muhammad Faisal and Janjua, Sana Iqbal},
  journal={IEEE Access}, 
  title={Motion Sickness in Virtual Reality: An Empirical Evaluation}, 
  year={2020},
  volume={8},
  number={},
  pages={130486-130499},
  doi={10.1109/ACCESS.2020.3007076}}

\end{document}